\documentclass[letterpaper]{article}
\usepackage{aaai}
\usepackage{times}
\usepackage{helvet}
\usepackage{courier}

\usepackage{algorithm}
\usepackage{algorithmic}
\usepackage{url}
\usepackage{times,amsmath,amssymb,amsfonts,epsfig,graphicx}
\usepackage{amsthm}  
\usepackage{enumerate}
\usepackage{stackrel}
\usepackage{multirow}
\usepackage{subfigure}
\usepackage{psfrag}
\usepackage{color}
\usepackage{comment}
\usepackage{bbm}
\usepackage{booktabs} 
\usepackage[normalem]{ulem}
\usepackage{xcolor}
\usepackage{framed}
\usepackage{tikz}
\usepackage{booktabs}
\usepackage{wrapfig}

\newcommand{\ra}[1]{\renewcommand{\arraystretch}{#1}}

\newcommand{\vectornorm}[1]{\|#1\|}

\newcommand{\abs}[1]{|#1|}
\newcommand{\dom}[1]{\mathrm{dom}(#1)}

\newcommand{\dimension}{n}
\newcommand{\numsam}{N}

\newcommand{\T}{{\boldsymbol \Theta}}
\newcommand{\x}{{\bf x}}

\newcommand{\decr}{\varepsilon}

\DeclareMathAlphabet{\mathbbb}{U}{bbold}{m}{n}

\DeclareMathOperator*{\argmin}{arg\,min}

\newtheorem{lemma}{Lemma}

\newtheorem{definition}{Definition}

\newtheorem{theorem}{Theorem}
\newtheorem{corollary}{Corollary}

 \def\independenT#1#2{\mathrel{\setbox0\hbox{$#1#2$}%
 \copy0\kern-\wd0\mkern4mu\box0}} 
 
\begin{document}
%
\title{Scalable sparse covariance estimation via self-concordance}
\author{Anastasios Kyrillidis, Rabeeh Karimi Mahabadi, Quoc Tran Dinh \and Volkan Cevher\thanks{This work is supported in part by the European Commission under Grant MIRG-268398, ERC Future Proof and SNF 200021-132548, SNF 200021-146750 and SNF CRSII2-147633.}\\
\'Ecole Polytechnique F\'ed\'erale de Lausanne\\
\texttt{\{anastasios.kyrillidis,rabeeh.karimimahabadi,quoc.trandinh,volkan.cevher\}}@epfl.ch}

\maketitle
\begin{abstract}
\begin{quote}
We consider the class of convex minimization problems, composed of a self-concordant function, such as the $\log\det$ metric, a convex data fidelity term $h(\cdot)$ and, a regularizing -- possibly non-smooth -- function $g(\cdot)$. 
This type of problems have recently attracted a great deal of interest, mainly due to their omnipresence in top-notch applications. Under this \emph{locally} Lipschitz continuous gradient setting, we analyze the convergence behavior of proximal Newton schemes with the added twist of a probable presence of inexact evaluations.
We prove attractive convergence rate guarantees and enhance state-of-the-art optimization schemes to accommodate such developments. 
Experimental results on sparse covariance estimation show the merits of our algorithm, both in terms of recovery efficiency and complexity. 
\end{quote}
\end{abstract}


\section{Introduction}
Convex $\ell_1$-regularized $\log\det$ divergence criteria have been proven to produce -- both theoretically and empirically -- consistent modeling in diverse top-notch applications. The literature on the setup and utilization of such criteria is expanding with applications in Gaussian graphical learning \cite{dahl2008covariance,banerjee2008model,hsieh2011sparse}, sparse covariance estimation \cite{rothman2012positive}, Poisson-based imaging \cite{harmany2012spiral}, etc. 

In this paper, we focus on the sparse covariance estimation problem. Particularly, let $\left\{\x_j\right\}_{j=1}^{\numsam}$ be a collection of $\dimension$-variate random vectors, i.e., $\x_j \in \mathbb{R}^{\dimension} $, drawn from a joint probability distribution with covariance matrix ${\boldsymbol \Sigma}$. In this context, assume there may exist unknown marginal independences among the variables to discover; we note that $({\boldsymbol \Sigma})_{kl} = 0$ when the $k$-th and $l$-th variables are independent. Here, we assume ${\boldsymbol \Sigma}$ is unknown and {\it sparse}, i.e., only a small number of entries are nonzero. Our goal is to recover the nonzero pattern of $\boldsymbol{\Sigma}$, as well as compute a good approximation, from a (possibly) limited sample corpus.

Mathematically, one way to approximate $\boldsymbol{\Sigma}$ is by solving: 
\begin{align}
\T^{\ast} = \argmin_{\T} \Big\{-\log\det(\T) + h(\T) + g(\T)\Big\}, \label{eq:1}
\end{align}
\noindent where $\T \in \mathbb{R}^{n \times n}$ is the optimization variable, $h(\cdot) := 1/(2\rho)\cdot\|\T - \widehat{\boldsymbol{\Sigma}}\|_\text{F}^2$ where $\widehat{\boldsymbol{\Sigma}}$ is the sample covariance and $g(\cdot) := \lambda/\rho \cdot \|\T\|_1$ is a convex nonsmooth regularizer function, accompanied with an easily computable proximity operator \cite{combettes2005signal}. 
and $\rho,~\lambda > 0$. 

Whereas there are several works \cite{becker2012quasi,lee2012proximal} that compute the minimizer of such composite objective functions, where the smooth term is generally a \emph{Lipschitz} continuous gradient function, in \eqref{eq:1} we consider a more tedious task: The objective function has only {\it locally Lipschitz} continuous gradient. However, one can easily observe that \eqref{eq:1} is \emph{self-concordant}; we refer to some notation and definitions in the Preliminaries section.  Within this context, \cite{tran2013composite} present a new convergence analysis and propose a series of proximal Newton schemes with provably quadratic convergence rate, under the assumption of \emph{exact algorithmic calculations at each step of the method}. 

Here, we extend the work of \cite{tran2013composite} to include \emph{inexact evaluations} and study how these errors propagate into the convergence rate. As a by-product, we apply these changes to propose the inexact \textbf{S}elf-\textbf{C}oncordant \textbf{OPT}imization (\texttt{iSCOPT}) framework. Finally, we consider the sparse covariance estimation problem as a running example for our discussions. The contributions are:
\begin{itemize}
\item [$(i)$] We consider locally Lipschitz continuous gradient convex problems, similar to \eqref{eq:1}, where errors are introduced in the calculation of the descent direction step. Our analysis indicates that inexact strategies achieve similar convergence rates as the corresponding exact ones. 
\item [$(ii)$] We present the inexact \texttt{SCOPT} solver (\texttt{iSCOPT}) for the sparse covariance estimation problem, with several variations that increase the convergence rate in practice. 
\end{itemize}


\section{Preliminaries}{\label{sec:prelim}}

\textbf{Notation:} 
We reserve $ {\bf vec} (\cdot) \colon \mathbb{R}^{n\times n} \to \mathbb{R}^{n^2 \times 1}$ to denote the vectorization operator which maps a matrix into a vector, by stacking its columns and, let ${\bf mat}(\cdot) \colon \mathbb{R}^{n^2 \times 1} \to \mathbb{R}^ {n \times n}$ be the inverse operation. $\mathbf{I}$ denotes the identity matrix. 


\begin{definition}[Self-concordant functions \cite{nesterov1994interior}] \label{concordant}
A convex function $\varphi(\cdot) : \dom{\varphi} \rightarrow \mathbb{R}$ is self-concordant if $\abs{\varphi'''(x)} \leq 2\varphi''(x)^{3/2}, \forall x \in \dom{\varphi}$.
A function $\psi(\cdot) : \dom{\psi} \rightarrow \mathbb{R}$ is self-concordant if $\varphi(t) \equiv \psi({\bf x}+t{\bf v})$ is self-concordant $\forall {\bf x} \in \dom{\psi}, {\bf v}\in \mathbb{R}^n$.
\end{definition}

For ${\bf v} \in \mathbb{R}^n$, we define $\vectornorm{{\bf v}}_{\bf x} \equiv \left( {\bf v}^T \nabla^2 f(\x) {\bf v} \right)^{1/2}$ as the local norm around ${\bf x} \in \dom{f}$  with respect to $f(\cdot)$. The corresponding dual norm is $\vectornorm{{\bf v}}_{\bf x}^{\ast} \equiv \max_{ \vectornorm{{\bf u}}_{\bf x} \leq 1} {\bf u}^T{\bf v} = \left( {\bf v}^T \nabla^2 f(\x)^{-1} {\bf v} \right)^{1/2}$. We define $\omega(\cdot) : \mathbb{R}\rightarrow\mathbb{R}_{+}$ as $\omega(t) \equiv t-\ln(1+t)$, and $\omega_{\ast}(\cdot) : [0,1] \rightarrow \mathbb{R}_{+}$ as $\omega_{\ast}(t) \equiv -t-\ln(1-t)$. Note that $\omega(\cdot)$ and $\omega_{\ast}(\cdot)$ are both nonnegative, strictly convex, and increasing. 

\noindent\textbf{Problem reformulation:} We can transform the matrix formulation of \eqref{eq:1} in the following vectorized problem:
\begin{equation}
\min_{\x \in \mathbb{R}^p} \Big\{F(\x): =\underbrace{-\log\det({\bf mat}(\x)) + h(\x)}_{=f(\x)} + g(\x) \Big\}, \label{eqn:prob_stat}
\end{equation} for ${\bf mat}(\x) \equiv \T,~\x \in \mathbb{R}^p,~p = n^2$, where $f(\x)$ is a convex, self-condordant function
and $g(\x)$ is a proper, lower semi-continuous and non-smooth convex regularization term. For our discussions, we assume $g(\x)$ is $\ell_1$-norm-based. 

\section{The algorithm in a nutshell}

For our convenience and without loss of generality, we use the vectorized reformulation in \eqref{eqn:prob_stat}. Here, we describe the \texttt{SCOPT} optimization framework, proposed in \cite{tran2013composite}. \texttt{SCOPT} generates a sequence of putative solutions $\lbrace \x_i \rbrace_{i \geq 0}$, according to:
\begin{align}{\label{eq:algo}}
\x_{i+1} = (1 - \tau_i)\x_{i} + \tau_i \boldsymbol{\delta}_i^{\star},\quad \tau_i = \frac{1}{\lambda_i + 1}, 
\end{align} where $\boldsymbol{\delta}_i^{\star} - \x_i\in \mathbb{R}^{p}$ is a descent direction, $\lambda_i := \|\boldsymbol{\delta}_i^{\star} - \x_i\|_{\x_i}$ and $\tau_i > 0$ is a step size along this direction. To compute $\boldsymbol{\delta}_i^{\star}$, we minimize the {\it non-smooth} convex surrogate of $F(\cdot)$ around $\x_i$; observe that $\lambda_i$ assumes \emph{exact} evaluations of $\boldsymbol{\delta}_i^\star$:
\begin{align}{\label{eq:surrogate1}}
\boldsymbol{\delta}_i^{\star} = \argmin_{\boldsymbol{\delta} \in \mathbb{R}^{p}} \left\{ U(\boldsymbol{\delta}, \x_i) + g(\boldsymbol{\delta}) \right\};
\end{align} $U(\boldsymbol{\delta}, \x_i)$ is a quadratic approximation of $f(\cdot)$ such that
$U(\boldsymbol{\delta}, \x_i) := f(\x_i) + \nabla f(\x_i)^T (\boldsymbol{\delta} - \x_i) + \frac{1}{2}(\boldsymbol{\delta} - \x_i)^T \nabla^2 f(\x_i) (\boldsymbol{\delta} - \x_i), $
\noindent where $\nabla f(\x_i)$ and $\nabla^2 f(\x_i) $ denote the gradient (first-order) and Hessian (second-order) information of function $f(\cdot)$ around $\x_i \in \dom{F}$, respectively.

While quadratic approximations of smooth functions (of the form $U(\boldsymbol{\delta}, \x_i)$) have become \emph{de facto} approaches for general convex \emph{smooth} objective functions, to the best of our knowledge, there are not many works considering a composite \emph{non-smooth and non-Lipschitz gradient} minimization case with provable convergence guarantees under the presence of errors in the descent direction evaluations.

%

\section{Inexact solutions in \eqref{eq:surrogate1}}

An important ingredient for our scheme is the calculation of the descent direction through \eqref{eq:surrogate1}. For sparsity based applications, we use FISTA -- a fast $\ell_1$-norm regularized gradient method for solving \eqref{eq:surrogate1}  \cite{beck2009fast} -- and describe how to efficiently implement such solver for the case of sparse covariance estimation where $f(\x) = \frac{1}{2\rho}\|\x - {\bf vec}(\widehat{\boldsymbol{\Sigma}})\|_2^2 -\log\det({\bf mat}(\x))$.

Given the current estimate $\x_i $, the gradient and the Hessian of $f(\cdot)$ around $\x_i$ can be computed respectively as: 
$\nabla f(\x_i)=\frac{1}{\rho} \left( \x_i - {\bf vec}(\widehat{\boldsymbol{\Sigma}})\right) - {\bf vec}\left({\bf mat}(\x_i)^{-1} \right) \in \mathbb{R}^{p \times 1}, $ 
$\nabla^2 f(\x_i) = \frac{\mathbf{I}}{\rho} + ({\bf mat}(\x_i)^{-1} \otimes {\bf mat}(\x_i)^{-1}) \in \mathbb{R}^{p \times p}. $ 
Given the above, let $ {\bf z} := \nabla f(\x_i) - \nabla^2 f(\x_i) \x_i $. After calculations on \eqref{eq:surrogate1}, we easily observe that \eqref{eq:surrogate1} is equivalent to:
\begin{align}
{\boldsymbol{\delta}}_i = \argmin_{\boldsymbol{\delta}} \Big\lbrace \underbrace{ \frac{1}{2}\boldsymbol{\delta}^T\nabla^2 f(\x_i)\boldsymbol{\delta}+{\bf z}^T \boldsymbol{\delta}}_{\varphi(\boldsymbol{\delta})}+g(\boldsymbol{\delta}) \Big\rbrace, \label{eqn:sub}
\end{align} where $\varphi(\cdot)$ is smooth and convex with Lipschitz constant $L$:
\begin{align}{\label{eq:lipschitz}}
L = \frac{1}{\rho} + \frac{1}{\lambda^2_{\min}( {\bf mat}(\x_i))},
\end{align} where $\lambda_{\min}(\cdot)$ denotes the minimum eigenvalue of a matrix. Combining the above quantities in a ISTA-like procedure \cite{daubechies2004iterative}, we have:
\begin{equation}
\boldsymbol{\delta}^{k+1} = \mathcal{S}_{\frac{\lambda}{L\rho}}\left(\boldsymbol{\delta}^k-\frac{1}{L}\nabla\varphi(\boldsymbol{\delta}^k)\right),
\end{equation} where we use superscript $k$ to denote the $k$-th iteration of the ISTA procedure (as opposed to the subscript $i$ for the $i$-th iteration of \eqref{eq:algo}). Here, $\nabla \varphi(\boldsymbol{\delta}^k)=  \nabla^2 f(\x_i)\boldsymbol{\delta}^k + \mathbf{z}$ and $\mathcal{S}_{\frac{\lambda}{L\rho}}(\x) := \mathrm{sign}(\x)\max\{\abs{\x} - \frac{\lambda}{L\rho}, 0\}$. Furthermore, to achieve an $\mathcal{O}(1/k^2)$ convergence rate, one can use acceleration techniques that lead to the FISTA algorithm, based on Nesterov's seminal work \cite{nesterov1983method}. We repeat and extend FISTA's guarantees, as described in the next theorem; the proof is provided in the supplementary material.

\begin{theorem}{\label{fista:theorem}}
Let $\{\boldsymbol{\delta}^k\}_{k \geq 1}$ be the sequence of estimates generated by FISTA. Moreover, define $G(\boldsymbol{\delta}) := U(\boldsymbol{\delta}, \x_i) + g(\boldsymbol{\delta})$ where $\boldsymbol{\delta}^{\star}$ is the minimizer with $\|\boldsymbol{\delta}^{\star}\|_2^2 \leq c$ for some global constant $c > 0$. Then, to achieve a solution $\boldsymbol{\delta}^K$ such that:
\begin{align}{\label{eq:FISTA_acc}}
G(\boldsymbol{\delta}^K) - G(\boldsymbol{\delta}^{\star}) \leq \epsilon, ~\epsilon > 0,
\end{align} the FISTA algorithm requires at least $K := \left\lceil \sqrt{\frac{2Lc}{\epsilon}} - 1 \right\rceil$ iterations. Moreover, it can be proved that:
\begin{align}
G(\boldsymbol{\delta}^K) - G(\boldsymbol{\delta}^{\star}) \geq \frac{1}{2} \|\boldsymbol{\delta}^{\star} - \boldsymbol{\delta}^K\|_{\x_i}^2 \nonumber
\end{align}
\end{theorem}

We note that, given accuracy $\epsilon$, $\boldsymbol{\delta}^K$ satisfies \eqref{eq:FISTA_acc} and $\boldsymbol{\delta}_i \leftarrow \boldsymbol{\delta}^K$ in the recursion \eqref{eq:algo}. In general, $c$ is not known apriori; in practice though, such a global constant can be found during execution, such that Theorem \ref{fista:theorem} is satisfied. A detailed description is given in the supplementary material.

For the sparse covariance problem, one can observe that $L$ and ${\bf z}$ are precomputed once before applying FISTA iterations. Given $\x_i$, we compute $\lambda_{\text{min}}( {\bf mat}(\x_i))$ in $\mathcal{O}(n^3)$ time complexity, while ${\bf z}$ can be computed with $\mathcal{O}(n^3)$ time cost using the Kronecker product property ${\bf vec}({\bf AXB}) = ({\bf B}^T \otimes {\bf A}){\bf vec}({\bf X})$. Similarly, $\nabla \varphi(\boldsymbol{\delta}^k)$ can be iteratively computed in $\mathcal{O}(n^3)$ time cost. Overall, the FISTA algorithm for this problem has $\mathcal{O}(K\cdot n^3)$ computational cost.

\section{\texttt{iSCOPT}: Inexact \texttt{SCOPT}}{\label{sec:algo}}

Assembling the ingredients described above leads to Algorithm \ref{algo:1}, which we call as the \textbf{I}next \textbf{S}elf-\textbf{C}oncordant \textbf{O}ptimization (\texttt{iSCOPT}) with the following convergence guarantees; our objective function satisfies the assumptions A.1, defined in \cite{tran2013composite}; the proof is provided in the supplementary material.

\begin{theorem}[Global convergence guarantee]\label{th:convergence}
Let $\tau_i := \frac{\decr_i - \sqrt{2\epsilon}}{\decr_i(\decr_i - \sqrt{2\epsilon} + 1)}  \in (0, 1)$ where $\decr_i := \|\boldsymbol{\delta}_i - \x_i\|_{\x_i}$ is the \emph{Newton decrement}, $\boldsymbol{\delta}_i$ is the solution of \eqref{eq:surrogate1} and $\epsilon$ is the requested accuracy for solving \eqref{eq:surrogate1}. Assume $\decr_i \geq \sqrt{2\epsilon}, ~\forall i,$ and let the set $\left\{\x \in \dom{F} : F(\x) \leq F(\x_0)\right\} $ be bounded. Then, \texttt{iSCOPT} generates $\lbrace \x_i \rbrace_{i \geq 0}$ such that $\x_{i+1}$ satisfies:
\begin{equation}
F(\x_{i+1}) \leq F(\x_i) - \xi(\tau_i), \quad \text{where}\nonumber
\end{equation} $\xi(\tau_i) = -\omega_{\ast}(\tau_i\decr_i) - \tau_i \left( \epsilon  -\frac{1}{2}\left(\decr_i - \sqrt{2\epsilon}\right)^2 - \frac{1}{2} \decr_i^2\right) \geq 0$, $\forall i$, i.e., $\lbrace F(\x_i)\rbrace_{i \geq 0}$ is a \emph{strictly non-increasing} sequence.
\end{theorem} 

\begin{figure*}[!htpb]
\begin{align}{\label{eq:thm5}}
\decr_{i+1} &\leq \frac{\left(1 - \tau_i \decr_i + \tau_i\sqrt{2\epsilon}\right)}{\left(1 - \tau_i \decr_i - \tau_i\sqrt{2\epsilon}\right)} \cdot \frac{\left(1 - \tau_i\left(1 - \sqrt{2\epsilon} \right) + \left(2\tau_i^2 - \tau_i\right)\decr_i\right)}{1 - 4\tau_i\left(\sqrt{2\epsilon} + \decr_i\right) + 2\tau_i^2\left(\sqrt{2\epsilon} + \decr_i\right)^2}\left(\sqrt{2\epsilon} + \decr_i\right)  + \sqrt{2\epsilon}
\end{align}
\hrule
\end{figure*}

\subsection{Quadratic convergence rate of \texttt{iSCOPT} algorithm}
For {\it strictly convex} criteria with unique solution $\x^{\star}$, the above proof guarantees convergence, i.e., $\lbrace \x_i \rbrace_{i \geq 0} \rightarrow \x^{\star}$ for sufficiently large $i$. Given this property, we prove the convergence rate towards the minimizer using {\it local information} in norm measures: as long as $\|\x_{i+1} - \x_{i}\|$ is away from $0$, the algorithm has not yet converged to $\x^{\star}$. On the other hand, as $\|\x_{i+1} - \x_{i} \| \rightarrow 0$, the sequence $\lbrace F(\x_i) \rbrace_{i \geq 0}$ converges to its minimum and $\lbrace \x_{i} \rbrace_{i \geq 0} \rightarrow \x^{\star}$, as $i $ increases.

\begin{algorithm}[!htpb]
\caption{Inexact \texttt{SCOPT} for sparse cov. estimation} 
\label{algo:1}
\begin{algorithmic}[1]
\STATE \textbf{Input:} $\x_0$, $\rho, \lambda > 0$, $\sigma = \frac{3}{40}$, $\epsilon, \gamma > 0$.
\STATE \textbf{while} $\varepsilon_i \leq \gamma$ or $i \leq I^{\text{max}}$ \textbf{ do}
\STATE ~~~Solve \eqref{eq:surrogate1} for $\boldsymbol{\delta}_i$ with accuracy $\epsilon$ and parameters $\rho, \lambda$.
\STATE ~~~Compute $\decr_i = \|\boldsymbol{\delta}_i - \x_i\|_{\x_i}$
\STATE ~~~\textbf{if} $(\decr_i > \sigma)$
\STATE ~~~~~~~$\x_{i+1} = (1 - \tau_i)\x_{i} + \tau_i \boldsymbol{\delta}_i$ for $\tau_i = \frac{\decr_i - \sqrt{2\epsilon}}{\decr_i(\decr_i - \sqrt{2\epsilon} + 1)} $.
\STATE ~~~\textbf{else} ~~~ $\x_{i+1} = \boldsymbol{\delta}_i$
\STATE \textbf{end while}
\end{algorithmic}
\end{algorithm} 

In our analysis, we use the weighted distance $\|\x_{i+1} - \x_i\|_{\x_i}$ to characterize the \emph{rate of convergence} of the putative solutions. By \eqref{eq:algo} and given $\boldsymbol{\delta}_i$ is a computable solution where $\|\boldsymbol{\delta}_i - \boldsymbol{\delta}_i^{\star}\|_{\x_i} \leq \sqrt{2\epsilon}$, we observe:
\begin{align}
\|\x_{i+1} - \x_i\|_{\x_i} = \|\tau_i \left(\boldsymbol{\delta}_i - \x_i\right)\|_{\x_i} \propto \|\boldsymbol{\delta}_i - \x_i \|_{\x_i} := \decr_i. \nonumber
\end{align} This setting is nearly algorithmic: given $\x_i$ and $\boldsymbol{\delta}_i$ at each iteration, we can observe the behavior of $\|\x_{i+1} - \x_i\|_{\x_i}$ through the {\it evolution} of $\lbrace \decr_i \rbrace_{i \geq 0}$ and identify the region where this sequence decreases with a quadratic rate. 

\begin{definition}
We define the quadratic convergence region $\mathcal{Q} = \lbrace \x_i: \x_i \in \text{\texttt{dom}}(F) \rbrace $ as such where $\lbrace \x_i \rbrace_{i \geq 0}$ satisfies $\decr_{i+1} \leq \beta \decr_{i}^2 + c$, for some constant $\beta > 0$, $\varepsilon_i < 1$ and bounded and small constant $c > 0$.
\end{definition} The following lemma provides a first step for a concrete characterization of $\beta$ for the \text{iSCOPT} algorithm; the proof can be found in the supplementary material.

\begin{lemma}\label{th:contraction}
For any $\tau_i$ selection, $\forall i$, the \texttt{iSCOPT} algorithm generates the sequence $\lbrace \decr_i \rbrace_{i \geq 0} $ such that \eqref{eq:thm5} holds.
\end{lemma}

We provide a series of corollaries and lemmata that justify the local quadratic convergence of our approach in theory. 
\begin{corollary}
In the ideal case where $\boldsymbol{\delta}_i^{\star}$ is computable exactly, i.e., $\epsilon = 0$, the \texttt{iSCOPT} algorithm is identical to the \texttt{SCOPT} algorithm \cite{tran2013composite}.
\end{corollary}

We apply the bound $\sqrt{2\epsilon} \leq \decr_i$ to simplify \eqref{eq:thm5} as:
\begin{align}{\label{eq:000}}
\decr_{i+1} \leq \frac{2}{1 - 2\tau_i \decr_i} \cdot \frac{1 - \tau_i + 2\tau_i^2 \decr_i}{1 - 8\tau_i\decr_i + 8\tau_i^2 \decr_i^2} \cdot \decr_i  + \sqrt{2\epsilon} 
\end{align}

Next, we describe the convergence rate of \texttt{iSCOPT} for the two distinct phases in our approach: full step size and damped step size; the proofs are provided in the supplementary material.
\begin{theorem}{\label{lemma:1}}
Assume $\tau_i = 1$. Then, \texttt{iSCOPT} satisfies:
\begin{align}
\decr_{i+1} \leq \beta \decr_i^2  + c, \nonumber
\end{align} where $\beta = \frac{4}{(1 - 2\decr_i)(1 - 8\decr_i + 8\decr_i^2)}  = \mathcal{O}\left(\frac{1}{1 - \decr_i}\right)$, $c = \sqrt{2\epsilon}$ and $\epsilon$ is user-defined. I.e., \texttt{iSCOPT} has \emph{locally quadratic convergence rate} where $c > 0$ is small-valued and bounded. Moreover, for $\decr_i \leq \frac{3}{40}, ~\forall i$, $\decr_{i+1} \leq 14 \decr_i^2  + \sqrt{2\epsilon}$.
\end{theorem}

\begin{theorem}{\label{lemma:2}}
Assume the damped-step case where $\tau_i = \frac{\decr_i - \sqrt{2\epsilon}}{\decr_i(\decr_i - \sqrt{2\epsilon} + 1)}  \in (0, 1)$. Then, \texttt{iSCOPT} satisfies:
\begin{align}
\decr_{i+1} \leq \beta \decr_i^2  + c, \nonumber
\end{align} where $\beta = \frac{2\left(\decr_i - \sqrt{2\epsilon} + 1\right)}{1 - 2\decr_i\left(\decr_i - \sqrt{2\epsilon}\right)} \cdot \frac{\decr_i^2 + 3\decr_i + 2\epsilon}{\left(2 + 6\sqrt{2\epsilon} + 2\epsilon\right) - \decr_i\left(6 + 2\sqrt{2\epsilon}\right)} = \mathcal{O}\left(\frac{1}{1 - \decr_i}\right)$ and $\epsilon$ is user-defined. I.e., \texttt{iSCOPT} has \emph{locally quadratic convergence rate} where $c > 0$ is small-valued and bounded. Moreover, for $\decr_i \leq \frac{3}{20}, ~\forall i$, $\decr_{i+1} \leq 14\decr_i^2  + \sqrt{2\epsilon}$.
\end{theorem}

\begin{table*}[!htpb]
\centering
\caption{\centering Summary of related work on sparse covariance estimation. Here, [1]: \cite{xue2012positive}, [2]: \cite{bien2011sparse}, [3]: \cite{rothman2012positive}, [4]: \cite{wang2012two}. All methods have the same $\mathcal{O}(n^3)$ time-complexity per iteration.} \label{tbl:related}
\ra{1.3}
\begin{small}
\begin{tabular}{l c c c c c c c c c c c c} \toprule
\multicolumn{1}{c}{} & \phantom{ab} & \multicolumn{1}{c}{[1]} &  \phantom{ab}  & \multicolumn{1}{c}{[2]} &  \phantom{ab}  & \multicolumn{1}{c}{[3]} &  \phantom{ab}  & \multicolumn{1}{c}{[4]} & \phantom{ab} & \multicolumn{1}{c}{This work}\\
\cmidrule{1-1} \cmidrule{3-3} \cmidrule{5-5} \cmidrule{7-7} \cmidrule{9-9} \cmidrule{11-11} 
\multicolumn{1}{c}{\# of tuning parameters} & & \multicolumn{1}{c}{2} & & \multicolumn{1}{c}{1}  & & \multicolumn{1}{c}{2} & & \multicolumn{1}{c}{1} & & \multicolumn{1}{c}{2}\\
\multicolumn{1}{c}{Convergence guarantee} & & \multicolumn{1}{c}{\checkmark} & & \multicolumn{1}{c}{\checkmark}  & & \multicolumn{1}{c}{\checkmark} & & \multicolumn{1}{c}{--} & & \multicolumn{1}{c}{\checkmark}\\
\multicolumn{1}{c}{Convergence rate} & & \multicolumn{1}{c}{Linear} & & \multicolumn{1}{c}{--} & & \multicolumn{1}{c}{--$^{\dagger}$} & & \multicolumn{1}{c}{--$^{\dagger}$} & & \multicolumn{1}{c}{Quadratic}\\
\multicolumn{1}{c}{Covariate distribution} & & \multicolumn{1}{c}{Any} & & \multicolumn{1}{c}{Gaussian} & & \multicolumn{1}{c}{Any} & & \multicolumn{1}{c}{Gaussian} & & \multicolumn{1}{c}{Any}\\
\bottomrule
\multicolumn{13}{l}{$^{\dagger}$To the best of our knowledge, block coordinate descent algorithms have known convergence \emph{only} for the case}\\ 
\multicolumn{13}{l}{of Lipschitz continuous gradient objective functions \cite{beck2013convergence}.}
\end{tabular}
\end{small}
\end{table*}

\subsection{An \texttt{iSCOPT} variant}
Starting from a point far away from the true solution, Newton-like methods might not show the expected convergence behavior. To tackle this issue, we can further perform Forward Line Search (FLS) \cite{tran2013composite}: starting from the current estimate $\tau_i$, one might perform a \emph{forward} binary search in the range $[\tau_i, 1]$. The selection of the new step size $\widehat{\tau}_i$ is taken as the maximum-valued step size in $[\tau_i, 1]$, as long as $\widehat{\tau}_i$ decreases the objective function $F(\cdot)$, while satisfying any constraints in the optimization. The supplementary material contains illustrative examples which we omit due to lack of space.


\section{Application to sparse covariance estimation}

Covariance estimation is an important problem, found in diverse research areas. In classic portfolio optimization \cite{markowitz1952portfolio}, the covariance matrix over the asset returns is unknown and even the estimation of the most significant dependencies among assets might lead to meaningful decisions for portfolio optimization. Other applications of the sparse covariance estimation include inference in gene dependency networks \cite{schafer2005empirical}, fMRI imaging \cite{varoquaux2010brain}, data mining \cite{alqallaf2002scalable}, etc. Overall, sparse covariance matrices come with nice properties such as natural graphical interpretation, whereas are easy to be transfered and stored.

To this end, we consider the following problem:

\noindent \textsc{Problem I:} {\it Given $ n $-dimensional samples $ \lbrace {\bf x}_j \rbrace_{j = 1}^{\numsam} $, drawn from a joint probability density function with {\em unknown} sparse covariance $\boldsymbol{\Sigma} \succ 0 $, we approximate $ \boldsymbol{\Sigma} $ as the solution to the following optimization problem for some $\lambda, \rho > 0 $:}
\begin{align}
\T^{\star} = \argmin_{\T} \Big\{\frac{1}{2\rho} \vectornorm{\T - \widehat{\boldsymbol{\Sigma}}}_F^2 - \log \det (\T) + \frac{\lambda}{\rho} \vectornorm{\T}_{1} \Big\} \nonumber 
\end{align} 

A summary of the related work on the sparse covariance problem is given in Table \ref{tbl:related} and a more detailed discussion is provided in the supplementary material.

\begin{table*}[!ht]
\centering
\caption{Summary of comparison results for time efficiency.} \label{tbl:syndata_table1}
\ra{1.3}
\begin{scriptsize}
\begin{tabular}{llll c ccc c ccc} \toprule
\multicolumn{4}{c}{Model} & \phantom{ab} & \multicolumn{3}{c}{$F(\T^\star)$ ($\times 10^2$)} & \phantom{ab}  & \multicolumn{3}{c}{Time (secs)} \\
\cmidrule{1-4} \cmidrule{6-8} \cmidrule{10-12}
 & \multicolumn{1}{c}{$n$} &  & \multicolumn{1}{c}{$\lambda$} & \phantom{ab} & [3] & \texttt{iSCOPT} & \texttt{iSCOPT} FLS & \phantom{ab} & [3] & \texttt{iSCOPT}  & \texttt{iSCOPT} FLS\\ \midrule
\multirow{6}{*}{$\boldsymbol{\Sigma}_3$} & \multirow{3}{*}{$100$} & $\frac{k}{n^2} = 0.05$ & $1$ &  & $32.013$ & $\textcolor[rgb]{0.4,0.1,0}{\mathbf{31.919}}$ & $\textcolor[rgb]{0.4,0.1,0}{\mathbf{31.919}}$ & &  $8.288$ & $9.996$ & $\textcolor[rgb]{0.4,0.1,0}{\mathbf{3.584}}$ \\
 &  & $\frac{k}{n^2} = 0.1$ & $0.5$ &  & $36.190$ & $\textcolor[rgb]{0.4,0.1,0}{\mathbf{34.689}}$ & $\textcolor[rgb]{0.4,0.1,0}{\mathbf{34.689}}$ & &  $10.470$ & $12.761$ & $\textcolor[rgb]{0.4,0.1,0}{\mathbf{5.012}}$ \\
 &  & $\frac{k}{n^2} = 0.2$ & $0.5$ &  & $62.143$ & $\textcolor[rgb]{0.4,0.1,0}{\mathbf{53.081}}$ & $\textcolor[rgb]{0.4,0.1,0}{\mathbf{53.081}}$ & &  $18.446$ & $14.720$ & $\textcolor[rgb]{0.4,0.1,0}{\mathbf{6.257}}$ \\
 \cmidrule{2-12}
& \multirow{3}{*}{$1000$} & $\frac{k}{n^2} = 0.05$ & $1$ &  & $-$ & $-$ & $\textcolor[rgb]{0.4,0.1,0}{\mathbf{2711.931}}$ & &  $> \texttt{T}$ & $> \texttt{T}$ & $\textcolor[rgb]{0.4,0.1,0}{\mathbf{759.724}}$ \\
 &  & $\frac{k}{n^2} = 0.1$ & $1$ &  & $-$ & $-$ & $\textcolor[rgb]{0.4,0.1,0}{\mathbf{4734.251}}$ & &  $> \texttt{T}$ & $> \texttt{T}$ & $\textcolor[rgb]{0.4,0.1,0}{\mathbf{875.344}}$ \\
 &  & $\frac{k}{n^2} = 0.2$ & $1$ &  & $-$ & $-$ & $\textcolor[rgb]{0.4,0.1,0}{\mathbf{5553.508}}$ & &  $> \texttt{T}$ & $> \texttt{T}$ & $\textcolor[rgb]{0.4,0.1,0}{\mathbf{1059.709}}$ \\
\bottomrule
\end{tabular}
\end{scriptsize}
\end{table*}

\begin{table*}[!ht]
\centering
\caption{Summary of comparison results for reconstruction of efficiency.} \label{tbl:syndata_table2}
\ra{1.3}
\begin{scriptsize}
\begin{tabular}{lll c ccc c ccc} \toprule
\multicolumn{3}{c}{Model} & \phantom{a} & \multicolumn{3}{c}{$\|\T^\star - \boldsymbol{\Sigma}\|_F/\|\boldsymbol{\Sigma}\|_F$} &  \phantom{a}  & \multicolumn{3}{c}{Time}  \\
\cmidrule{1-3} \cmidrule{5-7} \cmidrule{9-11} 
\multicolumn{1}{c}{} & \multicolumn{1}{c}{$n$} & \multicolumn{1}{c}{$N$} & \phantom{a} & [4] & [1] & \texttt{iSCOPT} FLS & \phantom{a}  & [4] & [1] & \texttt{iSCOPT} FLS \\ \midrule
\multirow{6}{*}{$\boldsymbol{\Sigma}_3$} & \multirow{3}{*}{$100$}  & $n/2$ & & $1.180$ & $0.912$ & $\textcolor[rgb]{0.4,0.1,0}{\mathbf{0.908}}$ & &  $0.456$ & $\textcolor[rgb]{0.4,0.1,0}{\mathbf{0.252}}$ & $2.604$\\
 &  & $n$ & & $0.920$ & $0.554$ & $\textcolor[rgb]{0.4,0.1,0}{\mathbf{0.542}}$ & & $0.494$ & $\textcolor[rgb]{0.4,0.1,0}{\mathbf{0.108}}$ & $0.155$ \\
 &  & $10n$ & & $0.396$ & $0.192$ & $\textcolor[rgb]{0.4,0.1,0}{\mathbf{0.190}}$ & & $0.451$ & $0.108$ & $\textcolor[rgb]{0.4,0.1,0}{\mathbf{0.054}}$  \\ \cmidrule{2-11}
 & \multirow{3}{*}{$2000$} & $n/2$ &  & $-$ & $\textcolor[rgb]{0.4,0.1,0}{\mathbf{0.428}}$ & $\textcolor[rgb]{0.4,0.1,0}{\mathbf{0.428}}$ & & $>\texttt{T}$ & $350.145$ & $\textcolor[rgb]{0.4,0.1,0}{\mathbf{203.515}}$  \\
 & & $n$ & & $-$ & $\textcolor[rgb]{0.4,0.1,0}{\mathbf{0.352}}$ & $\textcolor[rgb]{0.4,0.1,0}{\mathbf{0.352}}$ & & $>\texttt{T}$ & $385.340$ & $\textcolor[rgb]{0.4,0.1,0}{\mathbf{167.688}}$ \\
 & & $10n$ & & $-$ & $0.211$ & $\textcolor[rgb]{0.4,0.1,0}{\mathbf{0.209}}$ & & $>\texttt{T}$ & $401.970$ & $\textcolor[rgb]{0.4,0.1,0}{\mathbf{122.535}}$\\
\bottomrule
\end{tabular}
\end{scriptsize}
\end{table*}

\section{Experiments}{\label{sec:exp}}
All approaches are carefully implemented in \textsc{Matlab} code with no C-coded parts. In all cases, we set $I^{\text{max}} = 500, ~\gamma = 10^{-10}$ and $\epsilon = 10^{-8}$. A more extensive presentation of these results can be found in the supplementary material.

\subsection{Benchmarking \texttt{iSCOPT}: time efficiency}
To the best of our knowledge, only \cite{rothman2012positive} considers the same objective function as in \textsc{Problem I}. There, the proposed algorithm follows similar motions with the graphical Lasso method \cite{friedman2008sparse}. 

To show the merits of our approach as compared with the state-of-the-art in \cite{rothman2012positive}, we generate $\boldsymbol{\Sigma} \equiv \boldsymbol{\Sigma}_3$ as a random positive definite covariance matrix with $\|\boldsymbol{\Sigma}_3\|_0 = k$. In our experiments, we test sparsity levels $k$ such that $\frac{k}{n^2} = \lbrace 0.05, 0.1, 0.2 \rbrace$ and $n \in \lbrace 100, 1000, 2000\rbrace$. Without loss of generality, we assume that the variables are drawn from a joint Gaussian probability distribution. Given $\boldsymbol{\Sigma}$, we generate $\left\{\x_j\right\}_{j=1}^{\numsam}$ random $n$-variate vectors according to $\mathcal{N}(\mathbf{0}, \boldsymbol{\Sigma})$, where $\numsam = \frac{n}{2}$. Then, the sample covariance matrix $\widehat{\boldsymbol{\Sigma}} =\frac{1}{\numsam} \sum_{j = 1}^{\numsam} {\bf x}_j{\bf x}_j^T $ is ill-conditioned in all cases with $\text{rank}(\widehat{\boldsymbol{\Sigma}}) \leq \frac{n}{2}$. We observe that the number of unknowns is ${n \choose 2} = \frac{n(n-1)}{2}$; in our testbed, this corresponds to estimation of $4950$ up to $1,999,000$ variables. 
To compute $L$ in \eqref{eq:lipschitz}, we use a power method scheme with $\texttt{Pw} = 20$ iterations. All algorithms under comparison are initialized with $\x_0 = \textbf{vec}(\text{diag}(\widehat{\boldsymbol{\Sigma}}))$. As an execution wall time, we set $\texttt{T} = 3600$ seconds (1 hour). In all cases, we set $\rho = 0.1$. 

Table \ref{tbl:syndata_table1} contains the summary of results. Overall, the proposed framework shows superior performance across diverse configuration settings, both in terms of time complexity and objective function minimization efficiency: both \texttt{iSCOPT} and \texttt{iSCOPT} FLS find solutions with lower objective function value, as compared to \cite{rothman2012positive}, within the same time frame. The regular \texttt{iSCOPT} algorithm performs relatively well in terms of computational time as compared to the rest of the methods. However, its convergence rate heavily depends on the \emph{conservative} $\tau_i$ selection. We note that \eqref{eq:surrogate1} benefits from warm-start strategies that result in convergence in Step 3 of Algorithm \ref{algo:1} within a few steps. 

\subsection{Benchmarking \texttt{iSCOPT}: reconstruction efficiency}

We also measure the $\boldsymbol{\Sigma}$ reconstruction efficacy by solving \textsc{Problem I}, as compared to other optimization formulations for sparse covariance estimation. 
We compare our $\T^{\star}$ estimate with: $(i)$ the Alternating Direction Method of Multipliers (ADMM) implementation \cite{xue2012positive}, and $(ii)$ the coordinate descent algorithm \cite{wang2012two}. 

Table \ref{tbl:syndata_table2} aggregates the experimental results in terms of the normalized distance $\frac{\|\T^{\star} - \boldsymbol{\Sigma}\|_F}{\|\boldsymbol{\Sigma}\|_F}$ and the captured sparsity pattern in $\boldsymbol{\Sigma}$. Without loss of generality, we fix $\lambda = 0.5, \rho = 0.1$ for the case $n = 100$ and, $\lambda = 1.5, \rho = 0.1$ for the case $n = 2000$.
\texttt{iSCOPT} framework is at least as competitive with the state-of-the-art implementations for sparse covariance estimation. It is evident that the proposed \texttt{iSCOPT} variant, based on self-concordant analysis, is at least one order of magnitude faster than the rest of algorithms under comparison. In terms of reconstruction efficacy, using our proposed scheme, we can achieve marginally better $\boldsymbol{\Sigma}$ reconstruction performance, as compared to \cite{xue2012positive}. 

\begin{figure*}[t]
\begin{minipage}[c]{0.525\textwidth}
\centering
\includegraphics[width=1\textwidth]{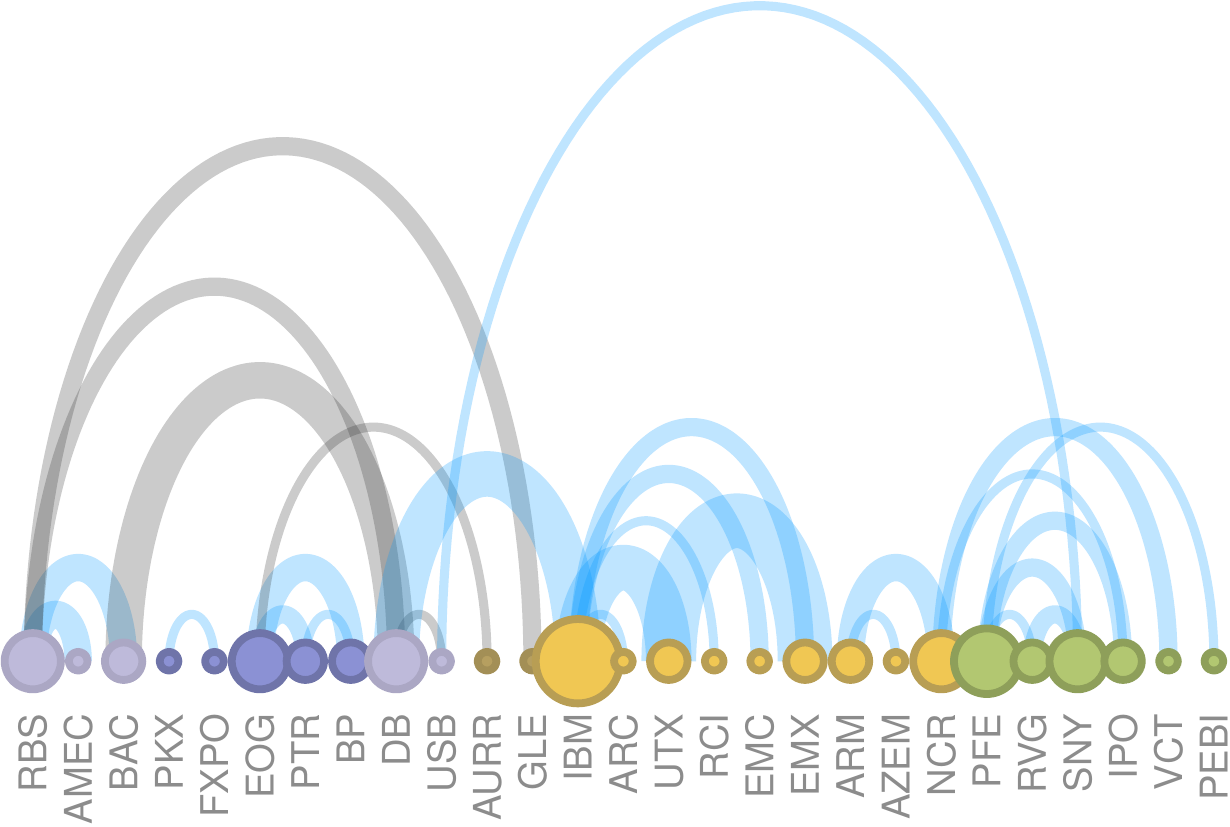} 
\label{fig:corr2}
\end{minipage} \hspace{0.15cm}
\begin{minipage}[c]{0.42\textwidth}
\begin{tiny}
\begin{tabular}{c c c c c c c} \toprule
\multicolumn{1}{c}{Stock Abbr.} & & \multicolumn{1}{c}{Company name}  &  & \multicolumn{1}{c}{Stock Abbr.} &  & \multicolumn{1}{c}{Company name}\\
\cmidrule{1-1} \cmidrule{3-3} \cmidrule{5-5} \cmidrule{7-7} 
\multicolumn{1}{c}{RBS} & & \multicolumn{1}{c}{Scotland Bank} & & \multicolumn{1}{c}{ARC} & & \multicolumn{1}{c}{Arc Document}\\ 
\multicolumn{1}{c}{AMEC} & & \multicolumn{1}{c}{AMEC Group} & & \multicolumn{1}{c}{UTX} & & \multicolumn{1}{c}{United Tech.} \\ 
\multicolumn{1}{c}{BAC} & & \multicolumn{1}{c}{Bank of America} & & \multicolumn{1}{c}{RCI} & & \multicolumn{1}{c}{Rogers Comm.}\\ 
\multicolumn{1}{c}{PKX} & & \multicolumn{1}{c}{Posco} & & \multicolumn{1}{c}{EMX} & & \multicolumn{1}{c}{EMX Industries}\\ 
\multicolumn{1}{c}{FXPO} & & \multicolumn{1}{c}{Ferrexpo} & & \multicolumn{1}{c}{ARM} & & \multicolumn{1}{c}{ARM Holdings} \\ 
\multicolumn{1}{c}{EOG} & & \multicolumn{1}{c}{EOG Resources} & & \multicolumn{1}{c}{AZEM} & & \multicolumn{1}{c}{Azem Chemicals}\\ 
\multicolumn{1}{c}{PTR} & & \multicolumn{1}{c}{PetroChina} & & \multicolumn{1}{c}{NCR} & & \multicolumn{1}{c}{NCR Electronics}\\
\multicolumn{1}{c}{BP} & & \multicolumn{1}{c}{BP} & & \multicolumn{1}{c}{PFE} & & \multicolumn{1}{c}{Pfizer Inc.}\\ 
\multicolumn{1}{c}{DB} & & \multicolumn{1}{c}{Deutsche Bank} & & \multicolumn{1}{c}{RVG} & & \multicolumn{1}{c}{Retro Virology} \\ 
\multicolumn{1}{c}{USB} & & \multicolumn{1}{c}{U.S. Bank Corp.} & & \multicolumn{1}{c}{SNY} & & \multicolumn{1}{c}{Sanofi health} \\ 
\multicolumn{1}{c}{AURR} & & \multicolumn{1}{c}{Aurora Russia} & & \multicolumn{1}{c}{IPO} & & \multicolumn{1}{c}{Intellectual Property}\\ 
\multicolumn{1}{c}{GLE} & & \multicolumn{1}{c}{Glencore} & &\multicolumn{1}{c}{VCT} & & \multicolumn{1}{c}{Victrex Chemicals}\\ 
\multicolumn{1}{c}{IBM} & & \multicolumn{1}{c}{IBM} & & \multicolumn{1}{c}{PEBI} & & \multicolumn{1}{c}{Port Erin BioFarma} \\ \bottomrule
\end{tabular}
\end{tiny}
\end{minipage}
\caption{We focus on three sectors: $(i)$ bank industry (light purple), $(ii)$ petroleum industry (dark purple), $(iii)$ Computer science and microelectronics industry (light yellow), $(iv)$ Pharmaceuticals/Chemistry industry (green). Any miscellaneous companies are denoted with dart yellow. Positive correlations are denoted with blue arcs; negative correlations with black arcs. The width of the arcs denotes the strength of the correlation - here, the maximum correlation (in magnitude) is $0.3934$. }
\label{fig:corr2}
\end{figure*}

\begin{table*}[tb]
\centering
\caption{Summary of portfolio optimization results-- all strategies considered achieve the requested return $\mu$.} \label{tbl:syndata_table3}
\ra{1.3}
\begin{footnotesize}
\begin{minipage}[c]{0.48\textwidth}
\begin{tabular}{lcc c ccc} \toprule
\multicolumn{3}{c}{Model} & \phantom{a} & \multicolumn{3}{c}{Risk $~~~\mathbf{w}^T \boldsymbol{\Sigma} \mathbf{w}$}  \\
\cmidrule{1-3} \cmidrule{5-7} 
\multicolumn{1}{c}{} & \multicolumn{1}{c}{$\lambda$} & \multicolumn{1}{c}{$\frac{k}{n^2}$ (\%)} & \phantom{a} & $\mathbf{w}(\widehat{\boldsymbol{\Sigma}})$ & $\mathbf{w}_{\text{equal}}$ & $\mathbf{w}(\T^{\star})$\\ \midrule
\multirow{9}{1.4cm}{$\boldsymbol{\Sigma}_3~~~~$ ($n = 1000$, $N = 90$)} & 1.4 & 0.5 & & 0.0760 & 0.0065 & $\textcolor[rgb]{0.4,0.1,0}{\mathbf{0.0053}}$ \\ 
& 1.7 & 1 & & 0.0810 & 0.0078 & $\textcolor[rgb]{0.4,0.1,0}{\mathbf{0.0059}}$ \\ 
& 2.3 & 5 & & 0.0902 & 0.0158 & $\textcolor[rgb]{0.4,0.1,0}{\mathbf{0.0129}}$ \\ 
& 2.7 & 7 & & 0.1968 & 0.0188 & $\textcolor[rgb]{0.4,0.1,0}{\mathbf{0.0159}}$ \\ 
& 3.0 & 10 & & 0.2232 & 0.0223 & $\textcolor[rgb]{0.4,0.1,0}{\mathbf{0.0196}}$ \\ 
& 3.8 & 15 & & 0.2463 & 0.0267 & $\textcolor[rgb]{0.4,0.1,0}{\mathbf{0.0231}}$ \\ 
& 4.5 & 20 & & 0.2408 & 0.0307 & $\textcolor[rgb]{0.4,0.1,0}{\mathbf{0.0257}}$ \\ 
& 4.5 & 30 & & 0.4925 & 0.0375 & $\textcolor[rgb]{0.4,0.1,0}{\mathbf{0.0365}}$ \\  \bottomrule
\end{tabular}
\end{minipage}
\hspace{0.3cm}
\begin{minipage}[c]{0.48\textwidth}
\begin{tabular}{lcc c ccc} \toprule
\multicolumn{3}{c}{Model} & \phantom{a} & \multicolumn{3}{c}{Risk $~~~\mathbf{w}^T \boldsymbol{\Sigma} \mathbf{w}$}  \\
\cmidrule{1-3} \cmidrule{5-7} 
\multicolumn{1}{c}{} & \multicolumn{1}{c}{$\lambda$} & \multicolumn{1}{c}{$\frac{k}{n^2}$ (\%)} & \phantom{a} & $\mathbf{w}(\widehat{\boldsymbol{\Sigma}})$ & $\mathbf{w}_{\text{equal}}$ & $\mathbf{w}(\T^{\star})$\\ \midrule
\multirow{9}{1.4cm}{$\boldsymbol{\Sigma}_3~~~~$ ($n = 1000$, $N = 180$)} & 1.4 & 0.5 & & 0.0223 & 0.0066 & $\textcolor[rgb]{0.4,0.1,0}{\mathbf{0.0050}}$ \\ 
& 1.7 & 1 & & 0.0233 & 0.0076 & $\textcolor[rgb]{0.4,0.1,0}{\mathbf{0.0072}}$ \\ 
& 2.3 & 5 & & 0.0513 & 0.0157 & $\textcolor[rgb]{0.4,0.1,0}{\mathbf{0.0115}}$ \\ 
& 2.7 & 7 & & 0.0529 & 0.0183 & $\textcolor[rgb]{0.4,0.1,0}{\mathbf{0.0139}}$ \\ 
& 3.0 & 10 & & 0.0706 & 0.0217 & $\textcolor[rgb]{0.4,0.1,0}{\mathbf{0.0177}}$ \\ 
& 3.8 & 15 & & 0.0876 & 0.0264 & $\textcolor[rgb]{0.4,0.1,0}{\mathbf{0.0202}}$ \\ 
& 4.5 & 20 & & 0.0872 & 0.0307 & $\textcolor[rgb]{0.4,0.1,0}{\mathbf{0.0227}}$ \\ 
& 4.5 & 30 & & 0.1075 & 0.0373 & $\textcolor[rgb]{0.4,0.1,0}{\mathbf{0.0291}}$ \\   \bottomrule
\end{tabular}
\end{minipage}
\end{footnotesize}
\end{table*}

\subsection{Sparse covariance estimates for portfolio optimization}
Classical mean-variance optimization (MVO) \cite{markowitz1952portfolio} corresponds to the following optimization problem:
\begin{equation}{\label{eq:mark}}
\begin{aligned}
& \underset{{\bf w}}{\text{minimize}}
& & {\bf w}^T \boldsymbol{\Sigma} {\bf w} \\
& \text{subject to}
& & {\bf w}^T{\bf r} = \mu, ~~\sum_i w_i = C, ~~w_i \geq 0,~\forall i.
\end{aligned}
\end{equation} Here, $\boldsymbol{\Sigma} \in \mathbb{S}_{+}^n$ is the {\it true} covariance matrix over a set of asset returns, ${\bf r} \in \mathbb{R}^n$ denotes the {\it true} asset returns of $n$ stocks, ${\bf w}$ represents a weighted probability distribution over the set of assets such that $\sum_i w_i = C$ and $C$ is the total capital to be invested. Without loss of generality, one can assume a normalized capital such that $\sum_i w_i = 1$. In such case, ${\bf w}^T \boldsymbol{\Sigma} {\bf w}$ is both the risk of the investment as well as a metric of {\it variance} of the portfolio selection.

In practice, both ${\bf r}$ and $\boldsymbol{\Sigma}$ are unknown and MVO requires an estimation for both. Empirical estimates, such as $\widehat{\boldsymbol{\Sigma}}$, quickly become problematic in the large scale: the data amount required increases quadratically to be commensurate with the degree of dimensionality. Due to such difficulties, even a simple {\it equal weighted portfolio} ${\bf w}$ such that $w_i = 1/n, ~\forall i$, is often preferred in practice \cite{demiguel2009optimal}. Nevertheless, practitioners assume that many elements of the covariance matrix are zero, a property which is appealing due to its interpretability and ease of estimation. Moreover, there are cases in practice where most of the variables are correlated to only a few others. 

Figure \ref{fig:corr2} shows some representative correlation estimates that we observed during the period $01.09.2009$ and $31.08.2013$. For this purpose, we use \texttt{iSCOPT} with $\lambda = 0.1$ and $\rho = 1$ to solve \textsc{Problem I} and sort the non-diagonal elements of $\T^{\star}$ to keep the most important correlations. We observe in practice some strong correlations between assets, while most of the rest entries in $\T^{\star}$ have significantly small magnitude. This dataset contains $2833$ stocks over a trading period of $1038$ days, crawled from the Yahoo Finance website\footnote{\url{http://finance.yahoo.com}}. Stocks are retrieved from stock markets in the America (e.g., Dow Jones, NYSE, etc.), Europe (e.g., London Stock Exchange, etc.), Asia (e.g., Nikkei, etc) and Africa (e.g., South Africa's exchange). 

\noindent \textbf{Out-of-sample performance with synthetic data:} It is apparent that both strong and weak correlations among stock assets are evident in practice. The behavior of non-diagonal entries in correlation matrix estimates is such that it is not easily distinguishable whether small values indicate weak dependence between variables or estimation fluctuations. Under these settings, \cite{hero2011large} propose that small values should be considered as zeros while only large values can be considered as good covariate estimates. 

To measure the performance of using a sparse covariance estimate in MVO, we assume the following synthetic case: Let $\boldsymbol{\Sigma} \succ 0$ be a synthetically generated Gaussian covariance matrix to represent the correlations among assets. Furthermore, assume that only $k$ entries of $\boldsymbol{\Sigma}$ are sufficiently larger than the rest of the entries.
In our experiments below we set $n = 1000$ and consider a time window of $N = 90, ~180$ days (i.e., a 3- and 6-month sampling period).

Given the above, both $\widehat{\boldsymbol{\Sigma}}$ and $\T^{\star}$ are calculated -- we use our algorithm for the latter. Using these two quantities, we then solve \eqref{eq:mark} for $\boldsymbol{\Sigma} \leftarrow \widehat{\boldsymbol{\Sigma}}$ and $\boldsymbol{\Sigma} \leftarrow \T^{\star}$ for various expected returns $\mu$ and record the computed minimum risk portfolios $\mathbf{w}(\widehat{\boldsymbol{\Sigma}})$ and $\mathbf{w}(\T^{\star})$, respectively. Finally, given $\mathbf{w}(\widehat{\boldsymbol{\Sigma}})$ and $\mathbf{w}(\T^{\star})$, as well as the  equal-weight portfolio $\mathbf{w}_{\text{equal}} := \frac{1}{n} \cdot \mathbbm{1}_{n \times 1}$, we report the risk/variances achieved by the constructed portfolios, using the ground truth covariance $\boldsymbol{\Sigma}$.
In Table \ref{tbl:syndata_table3}, we report lower variances $\mathbf{w}^T \boldsymbol{\Sigma} \mathbf{w}$ for portfolios $\mathbf{w}$ trained when $\T^{\star}$ is used in \eqref{eq:mark}, compared with the risk achieved by the equally-weighted portfolio or the sample covariance estimation where $\widehat{\boldsymbol{\Sigma}}$ is used. However, our approach comes with some cost to compute $\T^{\star}$. The empirical strategy with $\mathbf{w}(\widehat{\boldsymbol{\Sigma}})$ has the worst performance in terms of minimum risk achieved for most of our testings; we point out that, in this case, $\widehat{\boldsymbol{\Sigma}}$ is a rank-deficient positive semidefinite matrix.
\section{Discussion}
A drawback of our approach is the combined setup of the parameters $\lambda$ and $\rho$: one needs to identify selections that perform well on-the-fly, via a trial-and-error strategy. Unfortunately, such process might be inefficient in practice, especially in high dimensional cases. 
An interesting question to pursue is the \emph{adaptive} setup of at least one of $\lambda, ~\rho$. Such adaptive strategies have attracted a great deal of interest ; c.f., \cite{Hale2008}. 
One idea is to devise a path-following scheme with an adaptive $\rho$ selection, where the resulting scheme solves approximately a series of problems as $\rho \rightarrow 0$ is adaptively updated \cite{dinh2013inexact}. We hope this paper triggers future efforts towards this research direction for further investigation.

\bibliographystyle{aaai}
\bibliography{covbib}

\end{document}